%% file: main.tex
\documentclass[conference]{IEEEtran}
\usepackage{booktabs, afterpage}
\usepackage{amsmath}
\usepackage{graphicx}
\usepackage{pgfplots}
\usepackage{tikz}
\usetikzlibrary{shapes,arrows,arrows.meta}
\pgfplotsset{compat=1.13}
\usepackage{subcaption}
\usepackage{times}
\usepackage{xcolor}
\usepackage{soul}
\usepackage[utf8]{inputenc}
\newcommand*{\tran}{\top}
\pdfoutput = 1

\begin{document}
\title{ Deep Learning and Data Assimilation for Real-Time Production Prediction in Natural Gas Wells }


\author{\IEEEauthorblockN{Kelvin Loh}
\IEEEauthorblockA{\textit{Heat Transfer \& Fluid Dynamics} \\
\textit{TNO}\\
Delft, the Netherlands \\
kelvin.loh@tno.nl}
\and
\IEEEauthorblockN{Pejman Shoeibi Omrani}
\IEEEauthorblockA{\textit{Heat Transfer \& Fluid Dynamics} \\
\textit{TNO}\\
Delft, the Netherlands \\
pejman.shoeibiomrani@tno.nl}
\and
\IEEEauthorblockN{Ruud van der Linden}
\IEEEauthorblockA{\textit{Heat Transfer \& Fluid Dynamics} \\
\textit{TNO}\\
Delft, the Netherlands \\
ruud.vanderlinden@tno.nl}
}

\maketitle

\begin{abstract}
The prediction of the gas production from mature gas wells, due to their complex end-of-life behavior, is challenging and crucial for operational decision making. In this paper, we apply a modified deep LSTM model for prediction of the gas flow rates in mature gas wells, including the uncertainties in input parameters. Additionally, due to changes in the system in time and in order to increase the accuracy and robustness of the prediction, the Ensemble Kalman Filter (EnKF) is used to update the flow rate predictions based on new observations. The developed approach was tested on the data from two mature gas production wells in which their production is highly dynamic and suffering from salt deposition. The results show that the flow predictions using the EnKF updated model leads to better Jeffreys' J-divergences than the predictions without the EnKF model updating scheme.
\end{abstract}

\section{Introduction}
The mature North Sea gas wells are currently at their end-of-life, which make production predictions very challenging. With increasing energy demands, these predictions are increasingly crucial for operational decision making. Traditionally, expensive physics based models are used for prediction models, however, operators do have access to a large database of sensor data from these gas wells. It then becomes a natural extension that we explore the possibility of using a deep learning model to see if such methods can be applied to the field. We also would like to have a system which can be used in a "deploy and forget" manner, hence, the framework should be robust enough for different field conditions.

The oil and gas industry is very familiar with the use of Kalman filters as a data assimilation method, which is also used for model parameter estimation. We also see such work being done in ~\cite{Perez-Ortiz:2003:KFI:781392.781398}, where they used a Decoupled-Extended Kalman Filter (DEKF) for online training of a neural network model. Recently in computer vision, work has also been done to combine the two approaches together. It has been found that an LSTM Kalman Filter model for temporal regularization can outperform the standalone Kalman filter and standalone LSTM approaches \cite{DBLP:journals/corr/abs-1708-01885}. The extended Kalman filter (EKF) and its many variants including the DEKF does have a drawback in that it cannot handle highly nonlinear dynamical functions, therefore, in this paper, we would like to use the Ensemble Kalman Filter (EnKF) to update a prediction model parameter online. This is in the same manner as \cite{2017arXiv171208773C} which they call it the Bayesian LSTM. The difference being that we apply this approach to real valued time series data, while they used the results as an anomaly detection system. Reference \cite{mirikitani_4725078} has also observed that the EnKF trained RNNs outperform other EKF or gradient descent learning. Also widely used for online parameter estimation is the Unscented Kalman Filter (UKF), but for real valued time series data, it has been shown that the EnKF performs much better than the UKF \cite{hommels_2009}.

\subsection{Data Assimilation}
There exist a rich knowledge base for data assimilation methods within the weather prediction and reservoir engineering communities. In general, data assimilation methods can be split into two approaches, the deterministic approach which involves solving a minimization problem for all the data (3D/4D-VAR methods) and the probabilistic approach (which involves Bayesian inference/updating). The Kalman Filter belongs to the second group, and the EnKF and its many variants~\cite{Evensen2003} have proven to be very successful in atmospheric and oceanic sciences ~\cite{WRCR:WRCR13078}. A quick search of the many available work in EnKF itself citing \cite{Evensen2003} is already in the order of 2000s which gives an indication of how much wealth of knowledge these communities have.

\subsection{State of the art for deep learning models}
We restrict our overview for deep learning models to regression based models. As such, the most commonly used models for sequences and time series data are Recurrent Neural Networks (RNNs). In an overview of recent advances in RNNs,~\cite{2018arXiv180101078S} claimed that a well-trained RNN can model any dynamical system. Hence, the idea of using such models as the prediction model in our EnKF approach. Deep LSTMs and their variants were described by \cite{2013arXiv1303.5778G}  and \cite{Sak2014LongSM} which successfully modelled speech and acoustic modelling. Reference~\cite{DBLP:journals/corr/BianchiMKRJ17} investigated the use of RNNs for short term load predictions which being real-valued time series data, have the same characteristics as well production data.

\input{Test_case}

\input{Methodology}

\input{Results_discussion}

\section{Conclusions}
Given the objective of the paper and having tested the EnKF model updating approach on real data, we see from the results that this approach can help make the model predictions more robust. It shows that for the current application, the EnKF approach coupled with deep LSTMs can be invaluable when deployed in a real-time production optimization environment. Given the results obtained, an actual development on a real well monitoring system would be the next step.

\section*{Acknowledgments}
The authors would like to thank the Netherlands Enterprise Agency (RVO) and also wish to acknowledge the industry partners of the TKI Model Update consortium (Energie Beheer Nederland, Oranje-Nassau Energie, Total E \& P Nederland, Wintershall Noordzee) in alphabetical order, for the datasets and funding of this work.

\bibliographystyle{IEEEtran}

\end{document}

%% file: Test_case.tex

\section{Dataset}
\label{sec:Test_case}
\begin{figure}[htbp]
\centering
\includegraphics[width=0.45\textwidth]{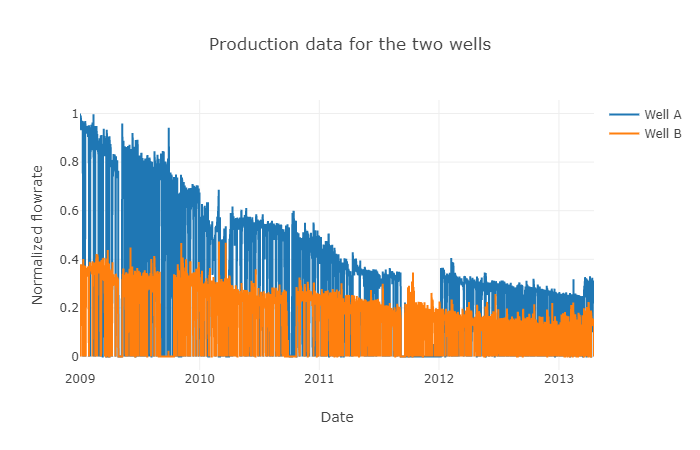}
\caption{Production data for both wells for the available dataset}
\label{fig:all_data_production}
\end{figure}

For this study, two mature gas wells located in the North sea were chosen. The production from these wells often suffer from salt precipitation due to the evaporation of the saline water. Since well production declines over time due to salt precipitation and ultimately may clog the well completely, it is important to predict the decline in the flow rate and plan shut-in and wash operations optimally. The production performance slowly decreases over time, due to depletion of the reservoir and irrecoverable salt precipitation in the near well bore. Thus, the forward model needs to be initially trained on the dataset to capture the production decline trend and afterwards being updated to take into account the changes in the system over time.
The production trends of these two wells are shown in Fig.~\ref{fig:all_data_production}. The data from these wells is available from Jan 01, 2009 to April 18, 2013. The data consists of pressure, temperature, flow rate and top-side choke valve opening in 10 minute intervals.

\subsection{Normalization}
\label{subsec:Dataset}
For all the variables in the dataset, we shift and scale the data using the minimum and range of the data from Well A respectively. They are each assumed to be Gaussian in their distributions. The six normalized input variables are given in Table~\ref{tab:variables} along with their scaled standard deviations. We obtain the scaled standard deviations by taking the claimed sensor accuracy (which represents a 95\% confidence interval) by the manufacturer, and multiply by the same scaling factor. This preserves the variance of the measurements when we sample from the normalized inputs distributions. 

\begin{table}[htbp]
    \centering
    \caption{Variable names and standard deviation}
    \begin{tabular}{ll} \toprule
       {Normalized Variable}  &  {Scaled Standard deviation} \\ \midrule
        Flow rate & 0.003\\
        Tubing Head Pressure Sensor 1 & 0.01\\
        Tubing Head Pressure Sensor 2 & 0.01\\
        Tubing Head Pressure Sensor 3 & 0.01\\
        Temperature & 0.04 \\ 
        Choke settings (valve opening) & 0 (Assumed perfect) \\ \bottomrule
    \end{tabular}
    \label{tab:variables}
\end{table}

\subsubsection{Training and validation}
\label{subsubsec:Training}
In this study we used the data of only Well A for the training, in the period from Jan 01, 2009 to Dec 31, 2011. Only the data of one well was used for the training justified by 
\begin{itemize}
    \item limited effort for training
    \item assess the usability of the trained model on one well and its prediction capability for other wells
\end{itemize}
The Well A data from Jan 01, 2012 to Apr 18, 2013 are used for the validation set.

\subsubsection{Testing Sets}
\label{subsubsec:Perf_Eval}
There are four different datasets being used to evaluate the performance of the developed approach relative to the baseline approach. We use two periods of data from the two wells. The first period is July 13-27, 2012, and the second period is March 10-24, 2013. The predictions from both approaches would be compared on these sets. The reason why the first period was chosen is because this set represents a typical cycle for well shutin and production decline profile for both wells. The second period was chosen because in this period, there are more shutins with shorter durations and the well dynamics involved can provide a challenging robustness test for both approaches.

%% file: Methodology.tex
\section{Methodology}
\label{sec:Methodology}
The summary for the methodology is that we use a modified deep LSTM model as our prediction model in the EnKF framework for parameter estimation. We then compare how close the prediction distributions of the EnKF updated model are to the measurements using Jeffreys' J-divergence. The baseline approach predictions are also compared against measurements using the same performance metric. We determine if the new approach of combining EnKF to update the bias parameter of this modified deep LSTM model, can improve and increase the robustness of the one timestep ahead predictions.

\subsection{Prediction Model}
\label{subsec:Prediction_model}
\begin{figure}[htbp]
\centering
\includegraphics[width=0.3\textwidth]{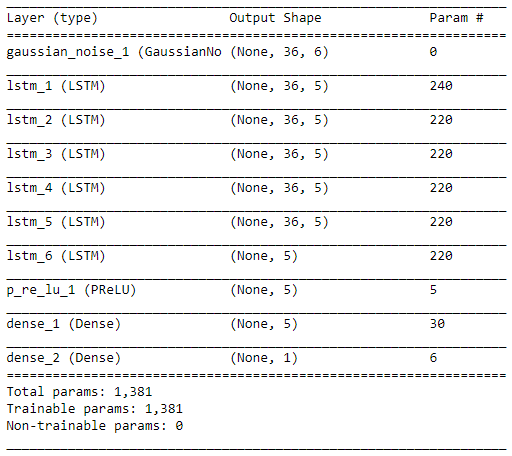}
\caption{Summary of model used for forward prediction}
\label{fig:forward_model}
\end{figure}

We use a modified deep LSTM network as described by \cite{Sak2014LongSM} for the prediction model. This allows us to capture more nonlinear dynamics at different time scales in the dataset, and we add a regularization input Gaussian noise layer with 0.1 standard deviation during training to prevent overfitting. Also, we add a parametric ReLU (PReLU) layer and two regular densely-connected neural network layers to the network. The reasoning behind using the PReLU and the 2 dense layers with the sigmoid activation function is to ensure that the flow rate predictions are always non-negative. Fig.~\ref{fig:forward_model} shows the summary for the model. Due to the limitations of space, we will not be discussing on the equations of an LSTM layer as it can be found in numerous literature already. The entire implementation of this model is based on the Keras framework using the Theano V0.9 backend ~\cite{chollet2015}. Training was performed using an Nvidia GTX 1080 GPU.

\begin{figure}[htbp]
\centering
\includegraphics[width=0.4\textwidth]{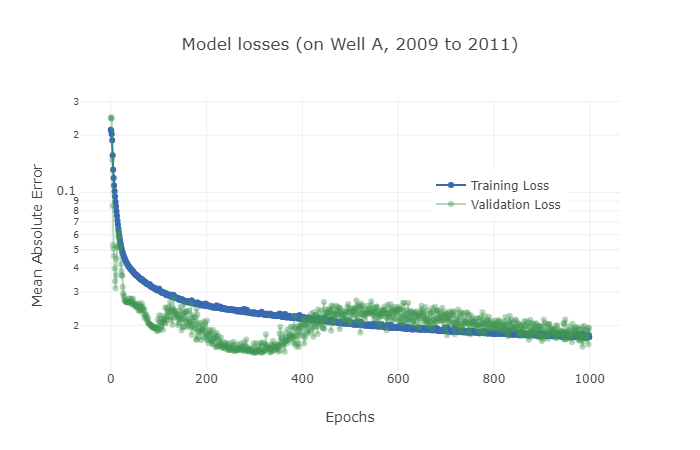}
\caption{Model losses during training}
\label{fig:model_losses}
\end{figure}

We use the Jan 01, 2009 to Dec 31, 2011 normalized dataset for Well A to train the deep LSTM model, see Section~\ref{subsec:Dataset}. Fig.~\ref{fig:model_losses} show the model losses during the training process as a function of epoch. We use the model weights at the last epoch for this paper.

All six variables defined in table~\ref{tab:variables} were used in the model training by a sliding window of 36 timesteps. In short, the prediction model can be represented in vectorized form by \eqref{eq:forward_equation}, where $n$ is the $n$-th timestep, $Q$ is the flow rate, $U$ is the choke valve settings, $\theta$ are the other four variables, and $W$ represents the modified deep LSTM model weights. Note that in \eqref{eq:forward_equation}, the choke valve settings are from timestep $n+1$ to $n-34$ instead of from $n$ to $n-35$. This is because the choke settings serve as the control parameter to the dynamical system.

\begin{align}
    \label{eq:forward_equation}
    Q_{n+1} &= f_{\text{D-LSTM}} \left( \begin{array}{c} Q_{n,n-1,...,n-35}\\ \theta_{n,n-1,...,n-35}\\ U_{n+1,n,...,n-34} \\ W \end{array} \right )
\end{align}

Note that all the variables in \eqref{eq:forward_equation} are considered to be random variables. We assume the weights of the model (except the bias parameter of the last layer) and choke valve settings to have 0 variances. For the baseline approach and to evaluate the Gaussianity of the model predictions, $Q_{n+1}$, we perform Monte Carlo simulations with 1000 samples using the normalized variables distributions.

\subsection{EnKF Model Updating Approach}
\label{subsec:enkf}
We define the system state, $x = [Q, W_{\text{bias}}]^\tran$, where additionally, $W_{\text{bias}}$ is the bias parameter of the last layer of the prediction model in section~\ref{subsec:Prediction_model}. This augmented system state is then used in the EnKF framework for bias and parameter estimation as described by \cite{Evensen2003}. Hence, the system state space effective size is only 2. The system dynamical model is given by \eqref{eq:space_equation}.

\begin{align}
    \label{eq:space_equation}
    x_{n+1} &= \left( \begin{array}{c} Q_{n+1} \\ W_{n+1} \end{array} \right) = \left( \begin{array}{c} f_{\text{D-LSTM}} (Q_n, \theta_n, U_{n+1}, W_n) \\ W_n \end{array} \right)
\end{align}

We visualize the EnKF approach with Fig.~\ref{fig:EnKF_block}.

\input{figure_enkf_tikz}

In Fig.~\ref{fig:EnKF_block}, we define $\mathbf{Q}_n$ as the model error covariance at timestep $n$, $P_{n+1}$ as the sample covariances, $K_n$ as the Kalman gain, $Z_n$ as the measurements vector, $R_n$ as the measurements covariance, and $M_n$ as the mapping from state space $X$ to measurements space $Z$. The red variable represents the prior distribution for the state $X$, which are used in the prediction performance comparisons with the baseline approach, and the orange variable represent the posterior distribution after measurements at that timestep has been assimilated.

We initialize the filter using the variable distributions in table~\ref{tab:variables} and $W_{\text{bias}} \sim \mathcal{N} ( W_\text{bias}^{\text{trained}}, 0.2 ) $. We also used ensemble sizes of 1000. We justify this by the fact that EnKF already works very well with sample sizes in the order of 100 with a dimension of $10^6$ \cite{2016arXiv160609321M}.

\subsection{Performance Metric}
\label{subsec:Comparison_metric}
We compare the performance of the predictions from the new EnKF model updating approach and the predictions from the baseline approach using Jeffreys' J-divergence ~\cite{Jeffreys453} for Gaussian distributions \eqref{eq:J_divergence}, where $P \sim \mathcal{N}(\mu_p,\sigma_p)$, $Q \sim \mathcal{N}(\mu_q,\sigma_q)$, and $D_{\text{KL}}$ is the Kullback-Leibler divergence.

\begin{align}
    \nonumber
    D_{\text{KL}} \left(P || Q \right) &= \log \left ( \frac{\sigma_q}{\sigma_p} \right ) + \frac{1}{2} \frac{\sigma_p^2 + \left( \mu_p - \mu_q \right)^2}{\sigma_q^2} - \frac{1}{2} \\
    \label{eq:J_divergence}
    D_{\text{J}} \left(P || Q \right) &= D_{\text{KL}} \left(P || Q \right) + D_{\text{KL}} \left(Q || P \right)
\end{align}

We always compare the predicted distributions with respect to the measurement distributions since we always assume the measurement to be the perfect distribution. Note that $D_{\text{J}} \left(P || Q \right) = D_{\text{J}} \left(Q || P \right)$.

%% file: figure_enkf_tikz.tex
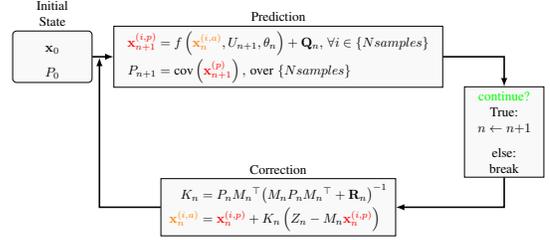
\begin{figure}[htbp]
    \centering
    \resizebox{.4\textwidth}{!}{\input{enkf_tikz}}
    \caption{Block diagram for the EnKF model update scheme}
    \label{fig:EnKF_block}
\end{figure}

%% file: enkf_tikz.tex
\tikzstyle{block} = [draw, rectangle, minimum width=6em, align=center,fill=gray!5]
\tikzstyle{arrow} = [-latex, very thick]
\begin{tikzpicture}[auto, node distance=2cm,>=latex']
    \setlength{\abovedisplayskip}{0pt}

    \node[text width=1cm,
          rounded corners=3pt,
          block,
          label={[above,align=center]{Initial\\State}}] at (-1, 0) (initial)
          {$$\mathbf{x}_0$$  $$P_0$$};
    \node at (0.25, 0.1) (sum) {};
    \node[block, text width=8.5cm,
          label={[above,align=center]{Prediction}}] at (5, 0) (prediction)
          {\begin{align*}
            {\color{red}\mathbf{x}_{n+1}^{(i,p)}} &= {\color{black} f} \left( {\color{orange} \mathbf{x}_n^{(i,a)}}, U_{n+1}, \theta_n \right)  + \mathbf{Q}_n, \, \forall i \in \{Nsamples\}\\
            P_{n+1} &= \text{cov} \left( {\color{red}\mathbf{x}^{(p)}_{n+1}} \right), \, \text{over $\{Nsamples\}$}
           \end{align*}};
    \node [block, right of=prediction,
            node distance=3cm, text width=1.4cm] at (8, -2) (iterUpdate)
            {{\color{green}continue?} True: $$n \leftarrow n + 1$$ else: break};
    \node [block, text width=6cm,
           label={[above,align=center]{Correction}}] at (5, -4) (correction)
           {\begin{align*}
              K_n &= P_n {\color{black}M_n}^\tran {\left ({\color{black}M_n} P_n {\color{black}M_n}^\tran + {\color{black}\mathbf{R}_n} \right)}^{-1}\\
              {\color{orange} \mathbf{x}_n^{(i,a)}} &= {\color{red}\mathbf{x}_{n}^{(i,p)}} + K_n \left( {\color{black}Z_n} - {\color{black}M_n} {\color{red}\mathbf{x}_{n}^{(i,p)}} \right)
            \end{align*}};

    \draw [arrow] (initial) -- (prediction);
    \draw [arrow] (prediction.east) -| (iterUpdate.north);
    \draw [arrow] (iterUpdate) |- (correction);
    \draw [arrow] (correction.west) -|  (sum);
\end{tikzpicture}

%% file: Results_discussion.tex
\section{Results and Discussion}
\label{sec:Results}
As mentioned in the methodology section (Sec~\ref{sec:Methodology}), we test the EnKF model update scheme on two wells, and two different production periods. In this section, we first determine if the predicted samples from the prediction model typically follow a Gaussian distribution. This is done by performing a normality test at each timestep in all the given dataset testing periods. After the test has been performed, this gives us some feel of how the EnKF would perform on the provided dataset. We then proceed to apply the EnKF algorithm on Well A for which the modified deep LSTM model had been trained on. After we determine the performance of the EnKF on the known well with both the production periods, we apply the same scheme on the second well, again with both the test production periods. Note that all of the uncertainty bounds shown in the figures within this section are the $2\sigma$ bounds which correspond to the usual 95\% confidence interval.

In terms of computational speed, all of the predictions using the EnKF model updated approach take a maximum of 0.35 wall clock-seconds per timestep for an ensemble size of 1000. Since the measurements are recorded at 10 minute intervals, it allows for real-time predictions and other automated applications on which we will not be discussing in this paper further.

\subsection{Testing the Gaussian assumption}
\label{subsec:test_gaussian}
The EnKF algorithm is optimal in the sense of Bayesian updating only if the system variables involved in the measurement update step are Gaussian distributed \cite{WRCR:WRCR13078}. Therefore, we need to check if the outputs from the forward model are Gaussian. We use the Shapiro-Wilk (SW) test \cite{shapiro1965} with $\alpha = 0.05$ as the threshold to reject the null hypothesis that the samples at a particular timestep is from a Gaussian distribution. We use this test because it has been shown that the SW test has a higher statistical power than any of the other three formal normality tests (Kolmogorov-Smirnov, Lilliefors, and Anderson-Darling) ~\cite{Razali2011PowerCO}, \cite{normality_tests}.

\begin{table}[htbp]
    \centering
    \caption{Well type and period of dataset with their associated number of timesteps which have rejected null hypothesis, out of a total of 1979 timesteps}
    \begin{tabular}{ccc} \toprule
       {Well ID}  &  {Period of dataset} & {Number of timesteps rejected} \\ \midrule
        A & Jul, 13-27 2012 & 208 \\
        A & Mar, 10-24 2012 & 303 \\
        B & Jul, 13-27 2012 & 469 \\
        B & Mar, 10-24 2012 & 1551 \\ \bottomrule
    \end{tabular}
    \label{tab:shapiro_reject}
\end{table}

\begin{figure}[htbp]
\centering
\includegraphics[width=0.45\textwidth]{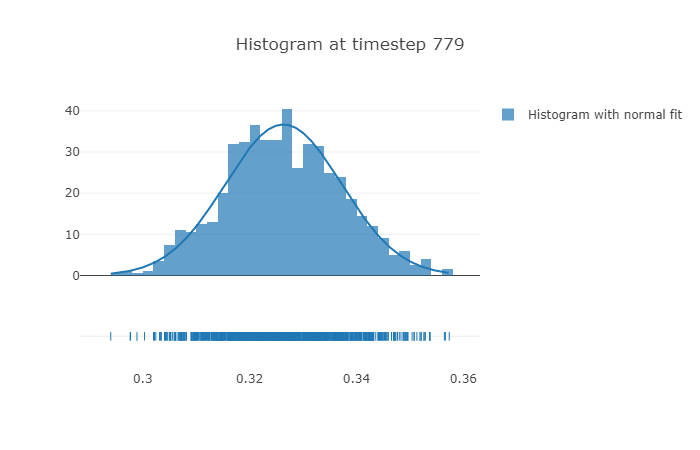}
\caption{Typical histogram of the forward model outputs for Well A in all the periods}
\label{fig:Well_A_histogram}
\end{figure}

\begin{figure}[htbp]
\centering
\includegraphics[width=0.45\textwidth]{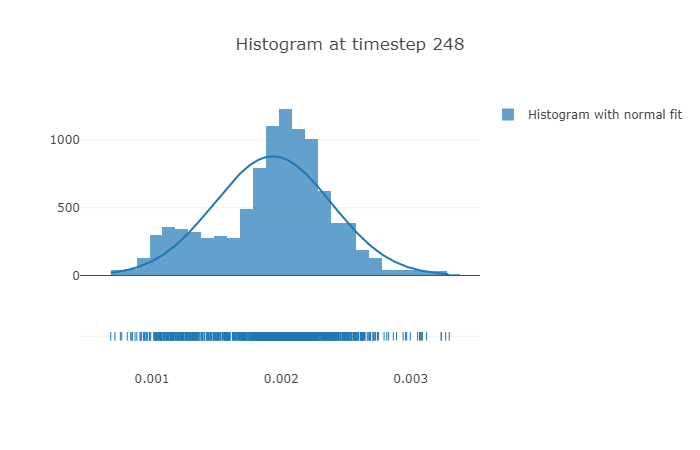}
\caption{Typical histogram of the forward model outputs for Well B in the March 10-24, 2013 dataset}
\label{fig:Well_B_histogram}
\end{figure}

Table~\ref{tab:shapiro_reject} shows the number of rejected null hypothesis timesteps for each of the different wells and dataset periods. We notice that the first three cases are predominantly Gaussian since only at most 469 out of 1979 timesteps failed the normality test. We now expect that the EnKF algorithm to perform optimally for the first three cases in the table, and perform suboptimally for the last case with Well B in the period of March 10-24, 2012. Fig.~\ref{fig:Well_A_histogram} shows a typical histogram shape of the predicted samples in each timestep for Well A in all of the periods. Contrast this to Fig.~\ref{fig:Well_B_histogram}, which shows a typical histogram of the predicted samples for Well B in the last case of Table~\ref{tab:shapiro_reject}. We can clearly see that the samples are not Gaussian distributed. The figures do confirm the results from Table~\ref{tab:shapiro_reject}.

\subsection{Testing on Well A with different periods}
\label{subsec:Well_A_results}
From the previous section (Sec~\ref{subsec:test_gaussian}), we would expect that the EnKF scheme to perform optimally for Well A. This is due to the fact that the prediction model was trained with data from this well. In particular, the many complex relationships between the physical variables of the well should be captured by the prediction model.

\begin{figure}[htbp]
\centering
\includegraphics[width=0.45\textwidth]{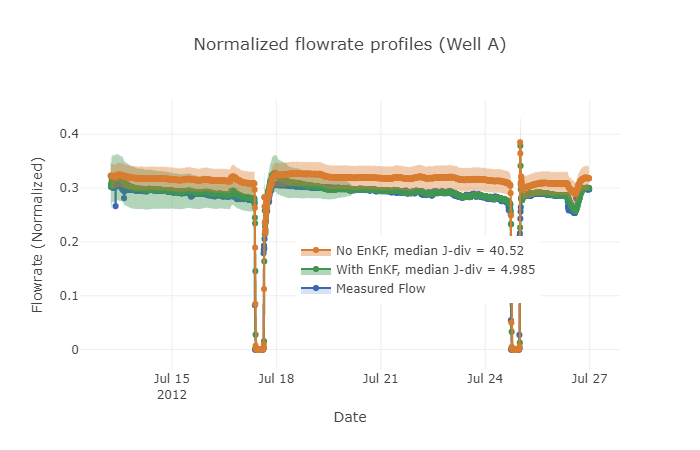}
\caption{Flow rate profiles of Well A for the period in July 13-27, 2012}
\label{fig:Well_A_profile_Jul2012}
\end{figure}

\begin{figure}[htbp]
\centering
\includegraphics[width=0.45\textwidth]{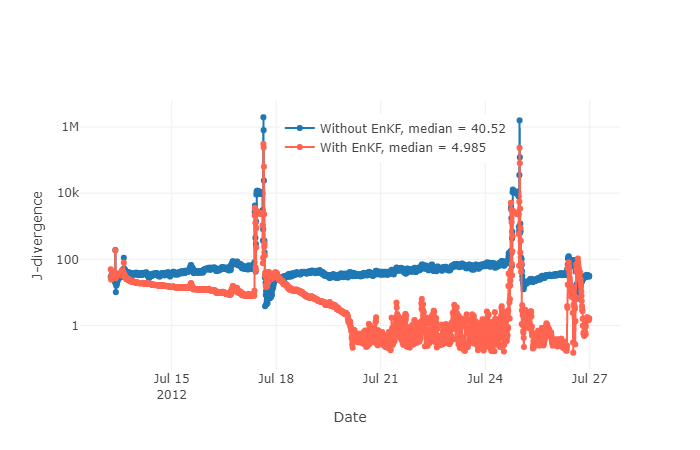}
\caption{J-divergence of Well A for the period in July 13-27, 2012}
\label{fig:Well_A_Jeff_Jul2012}
\end{figure}

Fig.~\ref{fig:Well_A_profile_Jul2012} show the predicted flow rate profiles between the baseline approach (No EnKF) and that of the EnKF model updating approach (With EnKF) for Well A during the July 13-27, 2012 period. It is seen that there are larger uncertainty bands around the mean of the No EnKF profile and with a higher median J-divergence of 40.52 against the median J-divergence of 4.985 of the EnKF updated profile. Still interesting to note is that the baseline algorithm does show the same physical relationships, which means that we do still see bumps in the predicted flowrates (between Jul 24-27) when the valve is opened again after a brief shutin (no flowrate) period. The only difference being that we see bias errors between the mean of the baseline case and that of the measurements. It then is not surprising that the EnKF updated model predictions would perform a lot better in the mean sense as well as in terms of matching the measured flow distributions indicated by a significantly lower median J-divergence. We can see that clearly with the J-divergence profiles for both these methods in Fig.~\ref{fig:Well_A_Jeff_Jul2012}. Ignoring the filter spin-up phase before July 21, we see that the J-divergence with the EnKF updated model is almost in the order of 1, and even with the shutin period, the filter is always able to follow the measured flowrate trajectory.

\begin{figure}[htbp]
\centering
\includegraphics[width=0.45\textwidth]{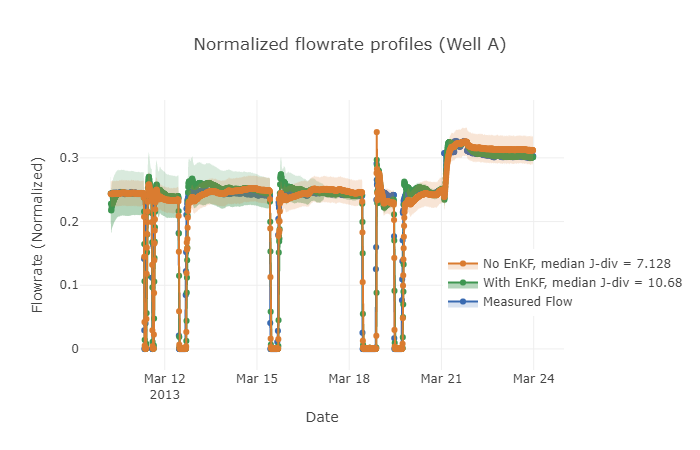}
\caption{Flow rate profiles of Well A for the period in Mar 10-24, 2013}
\label{fig:Well_A_profile_Mar2013}
\end{figure}

\begin{figure}[htbp]
\centering
\includegraphics[width=0.45\textwidth]{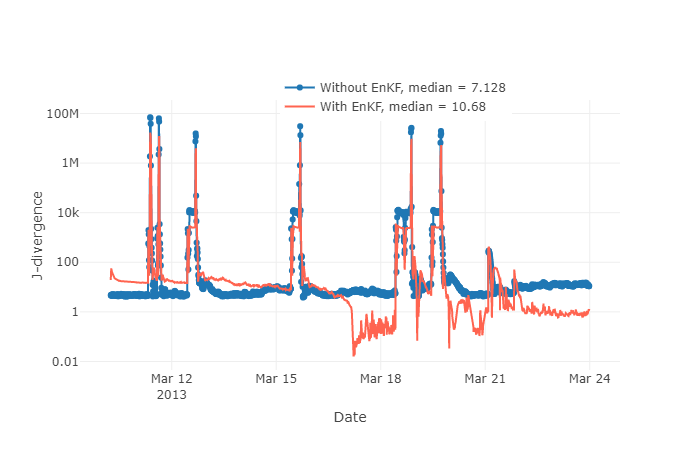}
\caption{J-divergence of Well A for the period in Mar 10-24, 2013}
\label{fig:Well_A_Jeff_Mar2013}
\end{figure}

The period of March 10-24, 2013 is also interesting because in this period, the baseline approach already performs quite well as can be seen from the mean and the J-divergence profiles shown in Fig.~\ref{fig:Well_A_profile_Mar2013}~\&~\ref{fig:Well_A_Jeff_Mar2013} respectively. We notice that the EnKF approach does perform slightly worse than the baseline approach with the median J-divergence for the baseline approach at 7.128 and for the EnKF updated approach at 10.68. This can again be attributed to the spin-up phase of the filter, as we see from Fig.~\ref{fig:Well_A_Jeff_Mar2013}, where the J-divergences of the EnKF updated approach are higher than the baseline approach up until Mar 16. After the spin-up phase, however, we notice that the J-divergences of the EnKF updated predictions are lower than the baseline predictions. Hence, we can conclude that for Well A, a deployment of the modified deep LSTM model with EnKF bias updating can provide accurate and robust predictions in a real-time manner.

\subsection{Testing on Well B with different periods}
\label{subsec:Well_B_results}
We have seen that the EnKF based approach did very well for Well A in both the testing periods. Trivially, this was due to the predictive model being trained on the dataset of Well A, and hence, any hidden relationships between the input variables would have been learned by the model during the training process. However, when we apply it to Well B, such hidden relationships might not hold anymore due to various physical factors affecting the well behaviour. We want to test if this new approach can be robust enough to automatically adapt to changing conditions when deployed on a totally unknown well by the prediction model. The dynamical behaviour should follow the same general trends (such as decreasing the valve opening should decrease the flow rate), but the bias parameter should be different due to the different well characteristics.

\begin{figure}[htbp]
\centering
\includegraphics[width=0.45\textwidth]{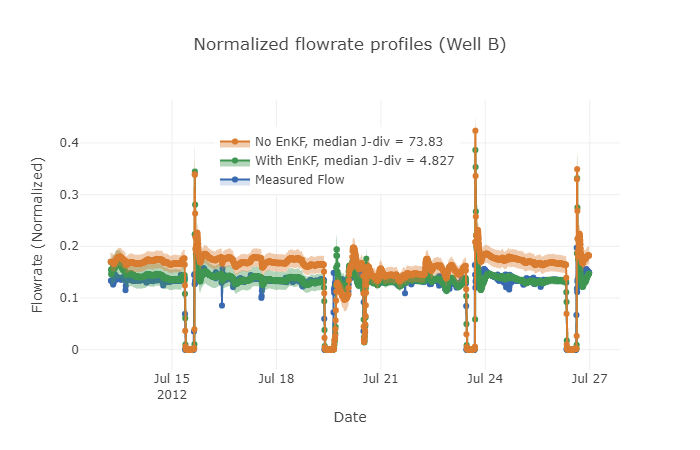}
\caption{Flow rate profiles of Well B for the period in July 13-27, 2012}
\label{fig:Well_B_profile_Jul2012}
\end{figure}

\begin{figure}[htbp]
\centering
\includegraphics[width=0.45\textwidth]{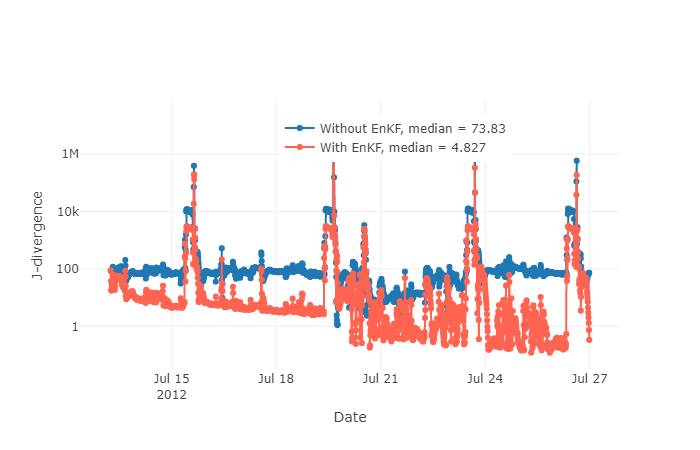}
\caption{J-divergence of Well B for the period in July 13-27, 2012}
\label{fig:Well_B_Jeff_Jul2012}
\end{figure}

We see from Fig.~\ref{fig:Well_B_profile_Jul2012}, the predicted production profiles between the two approaches and the measured production. The baseline approach still does quite well in the prediction but we see a mean bias between the predictions and the measurements. This indicates that we do indeed have model bias errors which can be corrected by the EnKF updated model approach. In the same figure, we observe that the bias error between the predictions of the EnKF updated model and the measurement has been minimized. Fig.~\ref{fig:Well_B_Jeff_Jul2012} provides a visual proof that indeed the developed approach does improve the prediction performance. The median J-divergence of the new approach is again lower than the baseline approach.

\begin{figure}[htbp]
\centering
\includegraphics[width=0.45\textwidth]{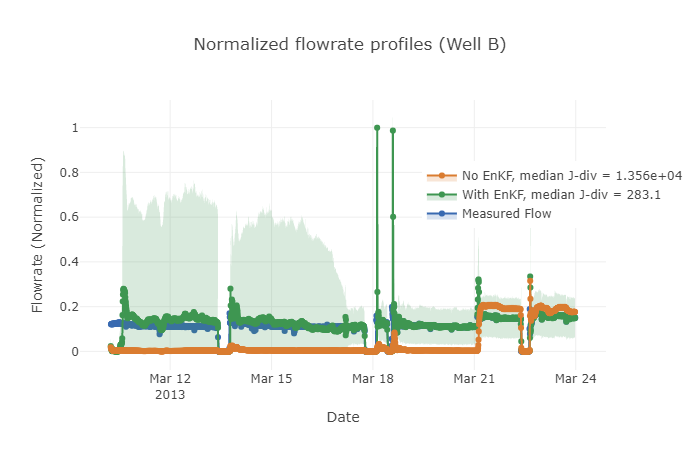}
\caption{Flow rate profiles of Well B for the period in Mar 10-24, 2013}
\label{fig:Well_B_profile_Mar2013}
\end{figure}

The Mar 10-24, 2013 period is the most challenging period for the two approaches. We can clearly see from Fig.~\ref{fig:Well_B_profile_Mar2013} that the baseline approach does not correctly predict the measured flow. There seems to be a high bias error in the prediction model at least between Mar 10 to Mar 21. Since we cannot inspect the well, we can only speculate that the structure of this well has managed to cause conditions similar as to the shut-in periods of Well A, which produces very low flow rate predictions. Despite such differences, we see that the EnKF updated model performs satisfactorily. We see that initially, the filter does struggle with very high uncertainties in the predictions especially during the spin-up period, but after it converges, it is able to track the measured profile correctly. In the mean sense, even the initial spin-up phase produces a vastly superior prediction over the baseline approach.

\begin{figure}[htbp]
\centering
\includegraphics[width=0.45\textwidth]{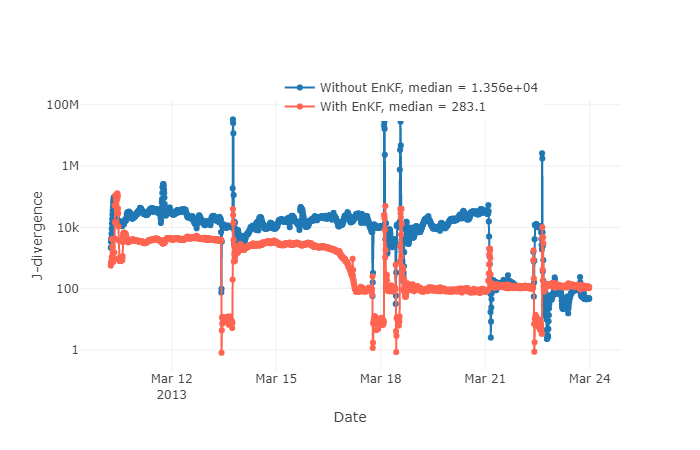}
\caption{J-divergence of Well B for the period in Mar 10-24, 2013}
\label{fig:Well_B_Jeff_Mar2013}
\end{figure}

We confirm that with the J-divergence profiles in Fig.~\ref{fig:Well_B_Jeff_Mar2013}. With a median J-divergence of 283.1, we do get a satisfactory prediction profile given the fact that the well conditions are not known to the model and we could only update the bias in the model parameter. This proves the utility of having an EnKF model updating approach by novice machine learning users and in real-time monitoring and optimization systems.

\begin{figure}[htbp]
\centering
\includegraphics[width=0.45\textwidth]{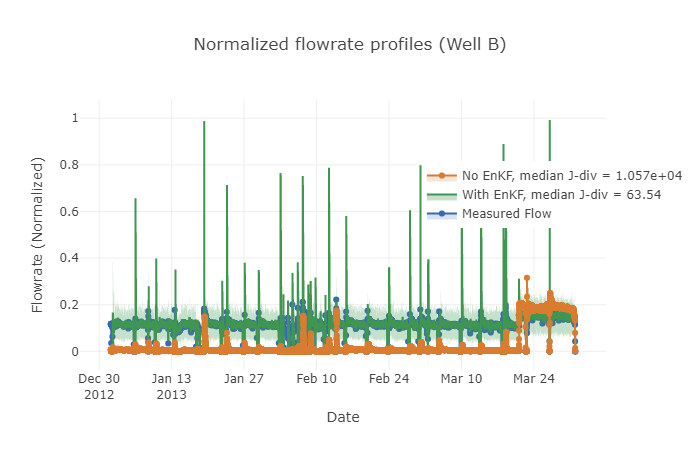}
\caption{Flow rate profiles of Well B for the period in Jan 01 - Mar 31, 2013}
\label{fig:Well_B_profile_full_Mar2013}
\end{figure}

\begin{figure}[htbp]
\centering
\includegraphics[width=0.45\textwidth]{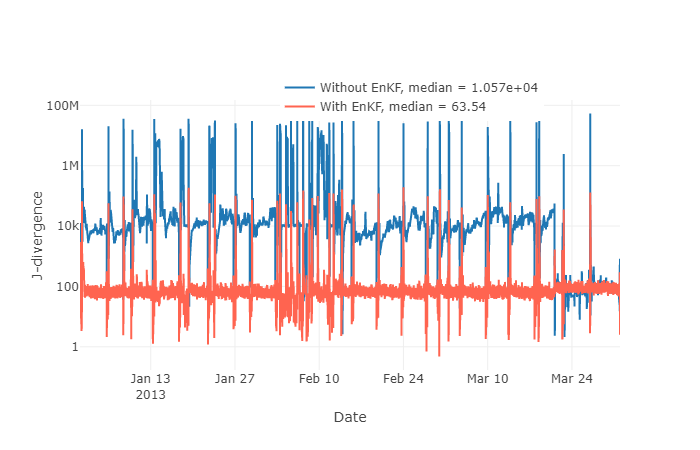}
\caption{J-divergence of Well B for the period in Jan 01 - Mar 31, 2013}
\label{fig:Well_B_jeff_full_Mar2013}
\end{figure}

For completeness sake, we also tested the approach on a longer period, to be more confident that the EnKF updated model approach can improve prediction robustness. This dataset is from Well B in the period of Jan 01 - Mar 31, 2013. The predicted production profiles are shown in Fig.~\ref{fig:Well_B_profile_full_Mar2013}, and the J-divergence profiles are shown in Fig.~\ref{fig:Well_B_jeff_full_Mar2013}. Based on the figures and the median J-divergence, it is indeed observed that the EnKF model updated approach can improve the well production predictions and adapt to different system conditions. We also notice that the filter spin-up period does not affect the long term performance of the predictions, hence, in a continuously running system, this spin-up behaviour will not be a major issue as long as we are aware of the existence and have methods to mitigate the behaviour~\cite{2008arXiv0806.0180K}.

